\documentclass{article}
\usepackage{preprint,times}

\usepackage{amsmath,amsfonts,bm}

\def\secref#1{section~\ref{#1}}

\def\eqref#1{equation~\ref{#1}}

\def\1{\bm{1}}

\DeclareMathAlphabet{\mathsfit}{\encodingdefault}{\sfdefault}{m}{sl}
\SetMathAlphabet{\mathsfit}{bold}{\encodingdefault}{\sfdefault}{bx}{n}

\usepackage{hyperref}
\usepackage{url}
\usepackage{booktabs}
\usepackage{enumitem}
\usepackage{graphicx}
\usepackage{subcaption}

\title{Telescopic Language Models}

\author{%
Zhilin Guo$^{1}$ \quad Boqiao Zhang$^{1}$ \quad Hakan Aktas$^{1}$ \quad Kyle Fogarty$^{1}$ \\
\bfseries Nursena Koprucu Aslan$^{1}$ \quad Wenzhao Li$^{1}$ \quad Canberk Baykal$^{1}$ \quad Albert Miao$^{1}$ \\
\bfseries Siyu Hong$^{1}$ \quad Yixiao Liu$^{2}$ \quad Adam Wu$^{1}$ \quad Ashish Kumar Singh$^{3}$ \\
\bfseries Sakar Khattar$^{3}$ \quad Chenliang Zhou$^{1}$ \quad Weihao Xia$^{1*}$ \quad Cristina Nader Vasconcelos$^{3}$ \\
\bfseries Cengiz Oztireli$^{1,3}$ \\[4pt]
$^{1}$University of Cambridge \qquad $^{2}$University of British Columbia \qquad $^{3}$Google \\
{\small $^{*}$Corresponding author: \texttt{wx258@cam.ac.uk}}
}

\ppfinalcopy %

\begin{document}

\maketitle
\lhead{}

\begin{abstract}
One deployed language model must often serve many compute budgets, yet
serving each budget still means a separate training or compression run per
point. We train a \emph{Telescopic Language Model} (TLM) to be that
continuum: a nested-capacity Transformer supervised by \emph{stochastic
prefix supervision with a full anchor}. At every step, one randomly
truncated prefix of the capacity axis is trained against the full
next-token target, alongside one full-capacity pass, so the trained
artifact is a valid language model at \emph{every} depth. Two
forward-backward passes per step, no architectural change, nothing extra
at inference. Fixed-exit suites such as Matryoshka Language Model Suites
(MLMS) occupy one point in this design space, and the point has a cost:
supervising only a few fixed exits leaves the nested model at chance level
everywhere else (perplexity $10^2$--$10^5$ in our baselines). On a 200M
proxy suite (20B FineWeb-Edu tokens, identical data stream for all
methods), a single TLM run is a valid language model at every one of its
twenty layer prefixes, in perplexity and on perplexity-sensitive
downstream tasks, reducing the area under the quality--budget curve by
43--44\% relative to the fixed-exit suites while matching them at full
capacity, at ${\sim}12\%$ lower GPU cost per run. The prefix sampling
density is a dial: concentrating it on a few depths recovers fixed-exit
quality there at the price of the continuum, so the operating points
become a training-time choice rather than an architectural one. These
results indicate that the training objective, not the nesting itself, is
what makes a model elastic.

\end{abstract}

\section{Introduction}
\label{sec:intro}
\suppressfloats[t]  %

\begin{figure}[t]
\centering
\includegraphics[width=0.85\columnwidth]{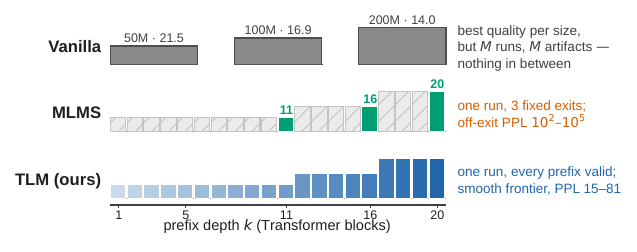}
\caption{One training run, a valid language model at every depth.
\emph{Top}: vanilla LLMs train a separate standalone model per budget (our
50M/100M/200M twins, annotated with validation perplexity) --- the best
quality per size, but $M$ runs, $M$ artifacts, and nothing in between.
\emph{Middle}: MLMS \citep{mlms2026} nests the budgets in one 20-layer
cascade but supervises only three fixed exits (green); every other prefix
is untrained and collapses (hatched). \emph{Bottom}: a Telescopic Language
Model trains one random prefix per step against the full target, so every
prefix of the same cascade is a valid language model (gradient: deeper
prefixes are larger models). The result is a continuum of operating points
from one run, with the full-size peak close to a separately trained twin
(14.99 vs.\ 13.98 perplexity).}
\label{fig:teaser}
\end{figure}

Language models are deployed across devices and latency budgets that differ
by orders of magnitude, from on-device assistants to datacenter endpoints.
Serving this range today means training a separate model per budget, or
compressing a large model post hoc and accepting the quality loss. Both
options multiply cost: either $M$ full training runs for $M$ operating
points, or a compression pipeline whose products trail models trained
natively at their size.

Nested-capacity cascades promise a cheaper route: sub-models of increasing
width and depth stacked into one Transformer trained end-to-end, so a single
run can in principle serve many sizes. The established instance, Matryoshka
Language Model Suites \citep{mlms2026}, supervises only $M{=}3$ fixed exits,
with online distillation from the largest sub-model. We find that the
objective, not the cascade, decides what gets served: at any depth other
than a trained exit, the nested model is effectively unusable, validation
perplexity rising from ${\sim}20$ at the exits to $10^2$--$10^5$ (up to
${\sim}4{\times}10^5$) at untrained depths (Figure~\ref{fig:frontier}). The
continuum of operating points that the cascade appears to offer is not
realized.

The continuum needs no new architecture, only a new objective. We train the
unchanged cascade with \emph{stochastic prefix supervision and a full
anchor}: at every optimization step, we sample one prefix length $k$ from a
distribution over all depths, evaluate the next-token loss of the
$k$-prefix against the full training target, and add the loss of the full
model on the same batch. The cost is two forward-backward passes per step;
the architecture and inference are unchanged: the trained artifact is
truncated at any depth and used directly. We call the resulting model a
\emph{Telescopic Language Model} (TLM), because its capacity extends and
collapses smoothly along a single ordered axis. The off-exit collapse, we
will show, is a property of the fixed-exit objective, not of the nesting.

On the 200M-parameter MLMS proxy suite, trained on 20B FineWeb-Edu tokens
with an identical data stream for every method, our main findings are:

\begin{itemize}[leftmargin=1.6em,itemsep=1pt,topsep=2pt]
\item \textbf{A continuum from one run.} Trained with a uniform prefix
sampler, TLM is a valid language model at every one of its twenty layer
prefixes. At depths MLMS does not train, where its perplexity reaches
$10^2$--$10^5$, TLM decreases smoothly (81 PPL at $k{=}1$ to 15 at
$k{=}20$); the area under the quality--budget curve drops by 43--44\%
(AULB 3.28 vs.\ 5.73--5.90).
\item \textbf{Valid below the smallest exit.} TLM serves ten operating
points below MLMS's 50M exit, down to $0.11\times$ full latency at PPL
26--81, where MLMS's own prefixes collapse.
\item \textbf{No added full-capacity tax.} The same uniform arm is on par
with the MLMS suites at the full 200M operating point (14.99 perplexity,
within their 14.96--15.42 span). Shifting the
sampler toward MLMS's exits trades the continuum for full-size quality: a
single seed reaches 14.65 (a 2.2\% edge over MLMS, within seed noise) and
cuts the remaining gap to a standalone vanilla twin by a third (4.8\% vs.\
7.1\%).
\item \textbf{A dial, not a wall.} The prefix sampling density $\pi$
controls the trade-off between prefix coverage and trained-exit quality:
concentrating $\pi$ on the MLMS exit depths improves quality at those
exits relative to uniform sampling but sacrifices off-exit coverage
(\S\ref{sec:ablations}). That endpoint shares MLMS's exit grid while
retaining TLM's sampled-prefix-plus-anchor objective --- the operating
points are a training-time choice, and no single setting dominates on both
axes.
\item \textbf{Lower cost per operating point.} The uniform TLM run costs
131 GPU-hours (${\sim}12\%$ less than a distillation-free MLMS suite at 149,
and $1.7\times$ less than the vanilla three-model suite at 216) while
delivering twenty operating points instead of three.
\end{itemize}

These results instantiate a broader training principle, stochastic prefix
supervision with a full anchor, previously validated on 3D Gaussian
splatting \citep{mgs2026}: when a model's capacity lives in an ordered list
of units, supervising one random prefix per step against the full target,
alongside the full model, turns that model into a continuum of models. We
show that on the language-model capacity axis this yields a single model
that \emph{is} a continuous quality--budget frontier --- every depth
served, the peak preserved relative to the fixed-exit suites --- and
characterize the trade-offs that govern where the sampler can be placed.

\section{Related Work}
\label{sec:related}

\textbf{Nested and Matryoshka models.} Matryoshka Representation Learning
\citep{kusupati2022mrl} trains embeddings that remain useful when truncated
to any of a fixed grid of dimensions, and nested dropout
\citep{rippel2014nested} showed that sampling a truncation point each step
induces an ordered, continuously truncatable representation. On the
language-model axis, MatFormer \citep{matformer2023} nests FFN widths inside
a decoder, and MLMS \citep{mlms2026} nests sub-models in both width and depth
with per-exit supervision plus distillation, delivering three operating
points per run. MatFormer's headline finding is that \emph{never-trained}
intermediate widths remain usable (Mix'n'Match), which sits at odds with our
motivating question: can one run deliver a valid model at every depth?
The reconciliation is where the nesting lives and what is supervised along
it: MatFormer nests width \emph{within} each layer under a single output head
trained on every step, so every truncation shares a calibrated readout,
whereas MLMS nests \emph{depth} across sub-models whose per-exit heads are
each trained only at their own exit, so an off-exit prefix pairs a trained
trunk with a head that never saw it (our readout-rescue control
attributes most of the collapse to this miscalibration). TLM trains the same cascade with a different objective: one
stochastically sampled prefix per step over a continuous depth grid, plus a
full-capacity anchor --- a continuum of operating points at a fixed two
passes per step, with fixed-exit supervision as the special case.

\textbf{Slimmable networks.} Slimmable and universally slimmable networks
\citep{yu2019slimmable,yu2019universallyslimmable} train CNNs that run at
several widths via a sandwich rule with in-place distillation; SortedNet
\citep{valipour2023sortednet} generalizes this to random sub-networks, and
Once-for-All \citep{cai2020onceforall} and Flextron \citep{cai2024flextron}
train or build elastic supernets that supervise multiple configurations per
step, with Once-for-All further distilling from the model's own full output;
we supervise exactly one sampled prefix plus the full model, and match the
full task target rather than a teacher distribution.

\textbf{Early exits and random depth.} Early-exit models
\citep{teerapittayanon2016branchynet,kaya2019shallowdeep,xin2020deebert,elhoushi2024layerskip}
attach heads at intermediate layers so easy inputs exit early,
Mixture-of-Depths \citep{raposo2024mixturedepths} varies compute per token,
and others learn per-token exit decisions
\citep{elbayad2020depthadaptive,schuster2022calm}; the exits remain a fixed
set of intended operating points. The closest training mechanism to ours is
random-depth supervision: stochastic depth \citep{huang2016stochasticdepth}
and LayerDrop \citep{fan2020layerdrop} drop blocks per step, and DynaBERT
\citep{hou2020dynabert} trains width- \emph{and} depth-adaptive sub-networks.
These share our ``train a random depth each step'' skeleton, but they drop
layers within a single shared-head stack rather than training one sampled
prefix of a width-growing cascade against the full task target, and they add
no full-capacity anchor; our LayerDrop control
tests whether random-depth robustness alone yields a continuum. Early exits
are complementary: a TLM prefix can itself serve as the draft in speculative
decoding \citep{specdec}.

\textbf{Other budget axes and post-hoc compression.} MQT-LLaVA
\citep{hu2024mqt} samples a token budget per step and MatryoshkaKV
\citep{lin2025matryoshkakv} samples KV-cache ranks, but neither combines a
sampled budget, a full anchor, and full-target matching on an ordered
capacity axis. Post-hoc compression --- structured pruning of width
\citep{ashkboos2024slicegpt,ma2023llmpruner} or depth
\citep{gromov2024unreasonable,men2024shortgpt} and per-size distillation
\citep{hinton2015distilling,sanh2019distilbert} --- derives smaller models
after the fact, one compression pass per target, and the products generally
lag a natively trained model of the same size; we compare against
independently trained vanilla models, the strongest in-hand competitor
(\S\ref{sec:experiments}).

\textbf{The same objective elsewhere.} Stochastic prefix supervision with a
full anchor was previously instantiated on 3D Gaussian splatting
\citep{mgs2026}, where the units are opacity-ordered splats and the same
two-pass objective yields a smooth, concave quality--budget frontier with
preserved peak quality. TLM carries the principle to language-model
capacity, where the units are nested Transformer blocks and per-budget
evaluation is expensive, which is precisely the regime in which the
two-pass design beats per-exit supervision on cost.

\section{Method}
\label{sec:method}

\subsection{Nested-capacity architecture}
\label{sec:method-arch}

TLM needs a nested-capacity cascade in which every prefix is a standalone
language model; we instantiate it with the Matryoshka Language Model Suite
(MLMS) architecture \citep{mlms2026}, used unchanged, and contribute the
training objective. Concretely, the cascade is an ordered stack of $M$
Llama-style
\citep{touvron2023llama} Transformer \citep{vaswani2017attention} sub-models
with strictly increasing widths
$D_1 < D_2 < \dots < D_M$ and depths $n_1, \dots, n_M$
($N = \sum_m n_m$ blocks in total). Each sub-model takes the previous
sub-model's output, concatenated at a \emph{junction} with a fresh slice of
the shared input embedding (the widened channels), and carries its own final
RMSNorm and LM head over a shared vocabulary, so every cascade prefix is a
standalone language model.

Formally, let $F_k(x;\theta)$ denote the next-token logits produced by the
\emph{$k$-prefix}: the first $k$ Transformer blocks, followed by the final
norm and LM head of the sub-model that contains block $k$. $F_N$ is the full
model. MLMS supervises only a fixed grid of exits
$k \in \{k_1, \dots, k_M\}$ (the block boundaries), minimizing
$\sum_m \mathcal{L}_{\mathrm{ce}}(F_{k_m}) +
\alpha_d \mathcal{L}_{\mathrm{distill}}$ at those exits; at any other depth
its prefixes are untrained and, as we show, collapse to $10^2$--$10^5$
perplexity.

\subsection{Stochastic prefix supervision with a full anchor}
\label{sec:method-objective}

TLM supervises a \emph{continuous} training signal over all depths, rather
than a fixed exit grid. Each optimization step performs exactly two forward-backward
passes over the same micro-batch $(x, y)$ of next-token pairs:
\begin{align}
\mathcal{L}_{\mathrm{step}}(\theta)
\;=\;
\underbrace{\lambda\; \ell\big(F_{\tilde{k}}(x;\theta),\, y\big)}_{\text{stochastic prefix pass},\ \tilde{k}\sim\pi}
\;+\;
\underbrace{\gamma\; \ell\big(F_{N}(x;\theta),\, y\big)}_{\text{full anchor pass}}
\label{eq:tlm-objective}
\end{align}
where $\ell$ is the token-level cross-entropy, $\pi$ is a sampling
distribution over prefix depths $\{1, \dots, N\}$, and $\tilde{k}\sim\pi$
is the \emph{sampled} prefix depth --- a random variable, resampled every
step, in contrast to $k$, which denotes a generic or evaluation depth. The
objective minimized is the expectation of this per-step loss,
$\mathcal{L}(\theta)=\mathbb{E}_{\tilde{k}\sim\pi}\,\mathcal{L}_{\mathrm{step}}(\theta)$.
Our default is
$\lambda = \gamma = 1$ with $\pi$ \emph{uniform} over depths, the only one
of the three samplers we study that trains every prefix (including
$k{=}1$) and the one that
yields a valid language model at all $N$ depths --- \emph{valid} in the
quantitative sense used throughout: perplexity finite and far below the
chance level of the vocabulary, decreasing smoothly with depth, on the
validation slice and on perplexity-sensitive downstream tasks. We also study a
\emph{log-uniform} sampler, $k=\lceil N^{u}\rceil$ with
$u\sim\mathrm{U}(0,1)$, which concentrates mass on small prefixes
($p(k)\approx 1/k$ for $k\ge2$, mean $k\approx6.9$, $k{=}1$ unsampled) and
an \emph{exit-grid} sampler concentrated on MLMS's exits
(\secref{sec:experiments}). The support is discrete --- the densest grid
the cascade admits, every integer depth --- so any budget is served
without retraining. The two passes act asymmetrically on the trunk: block
$j$ receives the anchor gradient every step but the prefix gradient only
when $\tilde{k}\ge j$, so $\pi$ induces a depth-dependent update schedule
in which shallow blocks update most often;
$\lambda{=}\gamma{=}1$ is the simplest weighting, and
\secref{sec:experiments} isolates each term. Nothing is added at
inference: the trained
model is truncated at any depth $k$ and used directly.

Each ingredient has a distinct role, and we isolate them by ablation in
\secref{sec:experiments}:

\textbf{Stochastic continuous budget.} Supervising a single randomly drawn
prefix per step (rather than all exits) decouples the number of supervised
depths from the per-step cost: we cover a continuum of $N$ depths for the
price of \emph{two} passes. MLMS instead applies a loss at every one of its
$M$ exits on every step (plus a distillation KL at each smaller exit), so its
per-step cost grows with the number of exits while still leaving all off-exit
depths untrained. Because $\tilde{k}$ is resampled every step, every prefix
receives an unbiased gradient signal over training. The supervision must be
end-to-end: readout calibration alone recovers only part of a collapsed
prefix (head-only repair remains ${\sim}6\times$ behind a trained prefix,
\secref{sec:experiments}).

\textbf{The full anchor.} The $F_N$ term ensures the largest
configuration receives a full-target gradient on every step, rather than
only on the fraction of steps that sample $k{=}N$. This is designed to
protect full-size quality (the no-anchor ablation in
\secref{sec:experiments} tests whether it is necessary), but it does not by
itself guarantee parity with an independently trained twin, because the
shared trunk is also shaped by the prefix gradients. The residual
$\Delta_{\mathrm{full}}$ is the price of that sharing
(\secref{sec:method-frontier}).

\textbf{Full-target matching.} The prefix pass is always trained against the
\emph{same} next-token target $y$ as the full model, never a cheaper or
reduced target, so every prefix is a valid language model for the same
task, not an auxiliary predictor.

\textbf{Optional distillation.} A third term,
$\alpha_d\, \mathcal{L}_{\mathrm{distill}}(F_{\tilde{k}} \leftarrow
\mathrm{sg}[F_N])$, can distill the full model's distribution into the
sampled prefix (stop-gradient teacher). We ablate it; the method does not
depend on it.

\subsection{The continuous quality--budget curve}
\label{sec:method-frontier}

The object of interest is the \emph{quality--budget curve} $Q(k)$:
validation perplexity
of the $k$-prefix as a function of depth (equivalently, of inference FLOPs,
which grow superlinearly in $k$ across the widening cascade; see Cost
accounting below). We evaluate $Q(k)$ on
every integer prefix $k = 1..N$ --- \emph{continuous} in coverage, the
densest grid the cascade admits; the smoothness of the curve is an
empirical finding, not part of the definition. We reserve
\emph{frontier} for the Pareto envelope derived from a method's valid
operating points (the LODA construction below). We summarize $Q(k)$ with:

\begin{itemize}[leftmargin=1.6em,itemsep=2pt,topsep=2pt]
\item \textbf{AULB} (area under the loss--budget curve, i.e.\ the
quality--budget curve $Q(k)$): the trapezoid
integral of token NLL over the integer prefix grid $k=1..N$, divided by the
grid span $N{-}1$ to return it to a mean-NLL scale (lower is better). AULB
collapses the whole frontier into one scalar for ablations. It weights each
unit of depth equally and is sensitive to the off-exit explosion under
fixed-exit objectives; we therefore treat the full per-prefix curves as the
primary evidence and compute FLOP-weighted, median, and deployment-range
variants as robustness checks (a large gap,
30--44\%, holds under all of them).
\item \textbf{Off-exit gap}: for a fixed-exit baseline (MLMS) evaluated on
the same grid, the per-depth difference
$Q_{\mathrm{MLMS}}(k) - Q_{\mathrm{TLM}}(k)$ at non-trained depths $k
\notin \{k_1, \dots, k_M\}$. This measures the continuum's added coverage
directly: quality at depths a fixed-exit suite does not train.
\item \textbf{$\Delta_{\mathrm{full}}$}: $Q(N)$ relative to an
independently trained vanilla twin of the full model. The full anchor is
designed to keep this small; any residual is the price of sharing the
trunk across prefixes.
\item \textbf{LODA} (level-of-detail area): the area under the
quality--\emph{throughput} frontier, the analog of
$\mathrm{AUC}_{\mathrm{fps}}$ in level-of-detail rendering
\citep{mgs2026}. A method's \emph{operating points} are its valid
configurations (every prefix for TLM; only the trained exits for MLMS, whose
off-exit prefixes collapse). The \emph{envelope} $\hat{q}(T)$ is the best
quality at throughput ${\ge}\,T$ --- a faster model can always be throttled,
so this is a max-from-right step function --- scoring the quality floor where
a method demonstrates nothing. Concretely: sort the method's valid operating
points by throughput and walk from the fastest point toward slower
throughputs; $\hat{q}(T)$
carries the best quality seen so far, holding each level as a flat step
until a slower point improves on it, so a suite with three exits contributes
three steps while a continuum contributes twenty. $\mathrm{LODA} =
\frac{100}{T_{\max}{-}T_{\min}}\int\hat{q}(T)\,dT$ (higher is better) is the
normalized area under this envelope. Where
AULB weights each unit of \emph{depth} equally, LODA weights each unit of
\emph{throughput} equally and credits only valid operating points, so it
measures the continuum's deployment value. We use chance-normalized benchmark
accuracy as the headline quality and clamped normalized perplexity as
robustness.
LODA is by construction a coverage metric --- it scores what a continuum is
for; the complementary per-exit comparison, where the suite retains an
edge, is reported alongside (\S\ref{sec:results}).
\end{itemize}

\paragraph{Cost accounting.} Training costs two forward-backward passes per
step regardless of how many depths are supervised, versus one shared cascade
pass with $M$ exit losses (plus a distillation teacher) for an $M$-exit MLMS
suite, and $M$ full independent trainings for a vanilla suite. Inference at
depth $k$ costs the cumulative FLOPs of the first $k$ blocks, sub-linear in
$k$ because block cost grows with width: the three exits $k{=}11/16/20$ use
${\sim}17\%/43\%/100\%$ of the full model's layer FLOPs. Unlike fixed-exit
suites, \emph{any} $k$ is available from a single training run.

\section{Experiments}
\label{sec:experiments}

\subsection{Experimental setup}
\label{sec:setup}

\textbf{Proxy suite.} All experiments use the 200M Matryoshka proxy suite of
\citet{mlms2026} (three nested widths: blocks 11@320 + 5@576 + 4@960,
head\_dim 64; \emph{norm} junctions; SmolLM2 tokenizer
\citep{allal2025smollm2}, vocab 49{,}152),
trained on FineWeb-Edu \citep{penedo2024fineweb} for 20B tokens per run:
packing at sequence length
2048, AdamW \citep{loshchilov2019adamw} $(\beta_1,\beta_2)=(0.9,0.95)$,
$\epsilon=10^{-8}$, peak lr
$4\times10^{-4}$, weight decay 0.01, grad clip 1.0, WSD schedule
\citep{hu2024minicpm} (1000 warmup
steps, 9.09\% cooldown), global batch $512\times2048$ tokens, bf16, seed 42.
Every run consumes the \emph{identical} data stream; the only independent
variable is the training objective.

\textbf{Methods compared.} (i) \textbf{MLMS}: our faithful reimplementation
of the four Table~3 configurations of \citet{mlms2026} (norm/nodistill,
plain, zero junction $\times$ distillation; validated against the published
numbers); the distillation arms use $\alpha_d{=}0.3$, a soft
cross-entropy to the stop-gradient largest-exit distribution at temperature
1, summed uniformly over the three exits (the full exit is pure
cross-entropy); (ii) \textbf{Vanilla twins}: standalone 50M/100M/200M
models trained on the same stream; (iii) \textbf{TLM}: stochastic prefix
supervision with full anchor ($\lambda=\gamma=1$, no distillation) under
three prefix samplers $\pi$: \emph{uniform} (our default; equal mass on all
twenty depths), \emph{log-uniform} (mass concentrated on small prefixes),
and \emph{exit-grid} (mass only on the three MLMS exit depths). The
log-uniform sampler draws $u\sim\mathrm{U}(0,1)$ and sets
$k=\lceil N^{u}\rceil$, so $p(k)\propto\log\frac{k}{k-1}\approx 1/k$ for
$k\ge2$ and $k{=}1$ has measure zero (mean $k\approx6.9$).

\textbf{Evaluation.} Held-out FineWeb-Edu validation perplexity; 7-benchmark
zero-shot accuracy (ARC-E/C, HellaSwag, LAMBADA, OpenBookQA, PIQA,
Winogrande; lm-eval-harness \citep{gao2024lmeval}, length-normalized
accuracy); OOD byte-PPL on
WikiText-103 \citep{merity2016pointer}, C4 \citep{dodge2021c4}, PG-19
\citep{rae2020compressive}, arXiv, and PubMed \citep{cohan2018discourse}; and
the quality--budget curve
$Q(k)$, the validation PPL of every integer prefix $k\in[1,20]$
(\S\ref{sec:method}). All PPL numbers for a given comparison use the same
held-out slice, so \emph{within-table} gaps are paired by construction
(identical eval data), which removes eval-set resampling variance;
finite-slice noise remains and is common across arms.

\subsection{Results \& analysis}
\label{sec:results}

\begin{figure}[t]
\centering
\includegraphics[width=\columnwidth]{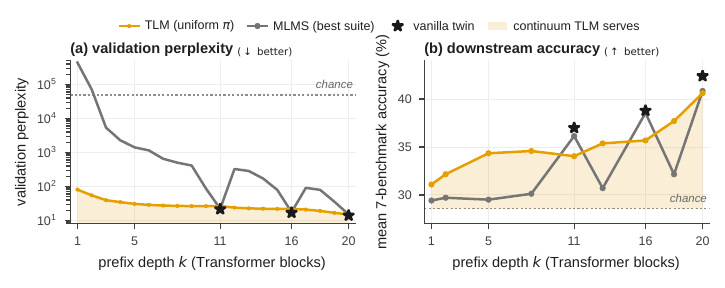}
\caption{\textbf{One training run, a valid model at every depth.} TLM
(amber) is a valid language model at every one of its twenty layer prefixes;
the shaded band is the continuum it serves. MLMS (grey) is valid only at its
three trained exits (dashed) and collapses elsewhere; at those exits it
remains stronger, the price of coverage (\S\ref{sec:results}). \textbf{(a)}
Validation perplexity of every prefix (log scale, lower is better; dotted:
chance): TLM
decreases smoothly across the whole range while MLMS falls to
$10^2$--$10^5$ between its exits. \textbf{(b)} Zero-shot downstream
accuracy (higher is better): TLM's continuum improves with depth while MLMS's off-exit
prefixes sit at chance; the band is TLM's above-chance quality at every
depth. Stars: independently trained vanilla twins (the per-size ceiling).
Area under the loss--budget frontier (AULB): TLM 3.28 vs.\ MLMS
5.73--5.90.}
\label{fig:frontier}
\end{figure}

\textbf{TLM is a valid model at every depth.}
Figure~\ref{fig:frontier} plots $Q(k)$ for TLM vs.\ MLMS. The uniform arm's
frontier is smooth and near-monotone throughout: it falls from 81 PPL at
$k{=}1$ to 15 at $k{=}20$, never spiking above 56 for any $k\ge2$. MLMS, by
contrast, is valid only at its three trained exits:
off-exit PPL reaches $10^2$--$10^5$ and beyond (e.g.\ $k{=}5$: 1{,}412;
$k{=}1$: 438{,}883), versus a chance level of 49{,}152. TLM cuts AULB from
5.73--5.90 (all four MLMS suites) to \textbf{3.28}, a 43--44\% reduction. The MLMS collapse
is partly a readout artifact: cheap head-only fine-tuning recovers it to 177
PPL at $k{=}5$ (suite AULB $5.90{\to}4.41$; Appendix~B),
but the rescued prefixes still lag TLM's trained prefixes by ${\sim}6\times$
and require per-prefix post-hoc tuning that TLM does not.

\begin{figure}[t]
\centering
\includegraphics[width=\columnwidth]{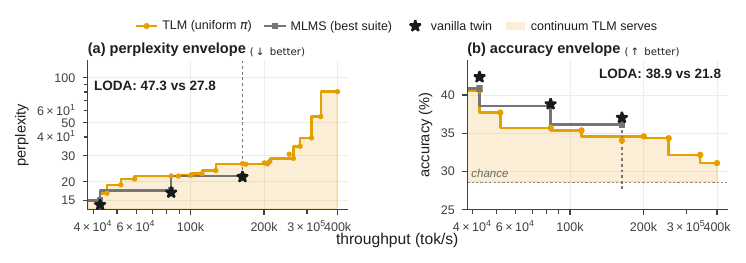}
\caption{\textbf{One TLM is a continuum of models; suites and vanilla
training yield isolated points.} Each marker is an operating point the method
actually trains. TLM-uniform (amber) serves \emph{every} prefill throughput
in the range; the shaded band is the continuum it delivers. MLMS (grey
squares) and vanilla twins (stars) exist only at their training sizes;
MLMS's exits edge TLM's prefixes at $k{=}11,16$ (the price of coverage) and
it collapses off-exit (dashes). Envelopes give the best quality served at
each throughput (log scale; faster rightward; a faster model can always be
throttled down). \textbf{(a)} Validation perplexity (log--log; lower is
better). \textbf{(b)} Mean 7-benchmark accuracy (higher is better); the band above the chance
floor (dotted) is exactly TLM's $\mathrm{LODA}_{\mathrm{acc}}$. LODA (area
under each envelope; perplexity uses a clamped normalised quality) is
annotated per method: TLM wins on both.}
\label{fig:loda}
\end{figure}

\textbf{TLM wins the deployment frontier.}
Table~\ref{tab:headline} gives the headline comparison and
Figure~\ref{fig:loda} the underlying quality--throughput frontier. LODA
(\S\ref{sec:method-frontier}) credits a method only at throughputs where it
serves a \emph{valid} model, so it measures what a continuum is \emph{for}.
The continuum arms win decisively: uniform $\pi$ reaches
$\mathrm{LODA}_{\mathrm{acc}}=38.9$ against MLMS's $21.8$ ($1.8\times$) and
$\mathrm{LODA}_{\mathrm{ppl}}=47.3$ against $27.8$. The win is structural:
below its smallest (50M) exit MLMS serves \emph{no} valid
model where TLM serves ten further points. The two non-continuum arms confirm the attribution: MLMS and
exit-grid $\pi$, which supervise only the exit depths, both collapse off-exit
and score far lower ($\mathrm{LODA}_{\mathrm{ppl}}$ $27.8$ and $25.7$) ---
dense stochastic supervision, not the architecture alone, creates the
deployment value.

\begin{table}[t]
\centering
\caption{Headline results (200M proxy suite, 20B tokens/run, identical data
stream). Full-size (200M) validation PPL; AULB over the prefix grid
$k{=}1..20$; $\mathrm{LODA}_{\mathrm{acc/ppl}}$, the area under the
quality--throughput frontier (\S\ref{sec:method-frontier}; higher is
better). \textbf{Bold} marks the best single-run cascade per column; the
vanilla twin is the per-size reference. `--': not evaluated.}
\label{tab:headline}
\small
\begin{tabular}{lcccc}
\toprule
Method & 200M PPL $\downarrow$ & AULB $\downarrow$ &
$\mathrm{LODA}_{\mathrm{acc}}$ $\uparrow$ &
$\mathrm{LODA}_{\mathrm{ppl}}$ $\uparrow$ \\
\midrule
MLMS (best suite)     & 14.98 & 5.90 & 21.8 & 27.8 \\
\midrule
TLM ($\pi$ uniform)    & 14.99 & \textbf{3.28} & \textbf{38.9} & 47.3 \\
TLM ($\pi$ log-unif.)  & 14.99 & 3.45 & 34.9 & \textbf{48.5} \\
TLM ($\pi$ exit-grid)  & \textbf{14.65} & 6.21 & -- & 25.7 \\
\midrule
vanilla twin (ref.)   & 13.98 & -- & -- & -- \\
\bottomrule
\end{tabular}
\end{table}

\textbf{The continuum is free at the peak.}
The uniform arm, the configuration that delivers the full continuum, is on
par with the MLMS suites at the full 200M operating point (14.99 perplexity,
within their 14.96--15.42 span, Appendix~A), so the
continuum costs nothing at the peak. The exit-grid arm goes further,
reaching 14.65 PPL with a comparable benchmark average (41.1 vs.\ 40.8),
though it abandons the continuum (\S\ref{sec:ablations}) and its sampler
also up-weights the full exit (the anchor every step plus the ${\sim}1/3$ of
prefix passes sampling $k{=}N$ give ${\sim}4/3\times$ MLMS's full-exit loss
weight).
Against the standalone vanilla twin (13.98), the full-size gap narrows from
MLMS's 7.1\% to TLM-exitgrid's 4.8\%, about a third. We read 14.99 vs.\ 14.98 as parity, and the 14.65
vs.\ 14.98 gap as \emph{consistent with} a
full-size edge rather than decisive (the suites remain single-seed); paired
second seeds of both TLM arms (Appendix~C) measure the
run-to-run spread directly, and the exit-grid edge replicates ($14.67$
vs.\ $14.65$).

\begin{figure}[t]
\centering
\includegraphics[width=\textwidth]{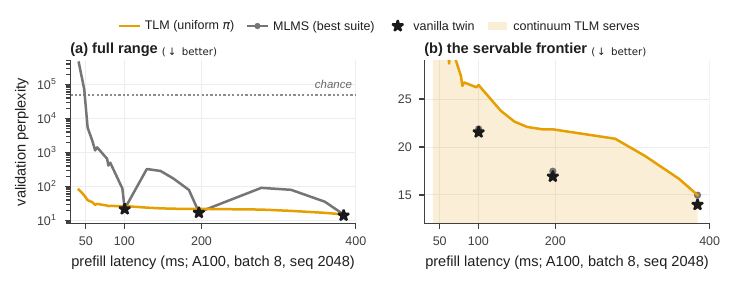}
\caption{\textbf{One run serves any latency; the suite serves three.}
Validation perplexity (lower is better) against measured prefill latency
(one A100, batch 8, sequence 2048, bf16). \emph{(a)} TLM-uniform (amber) is a valid language
model at every latency, while MLMS (grey) is servable only at its three
trained exits (circles) and collapses off-exit, rising to $10^2$--$10^5$
PPL up to the dotted chance line. \emph{(b)} Zoom on the servable range:
the band is the continuum TLM serves at every latency; MLMS's exits sit
below TLM at $k{=}11$ and $k{=}16$, its only valid points. Stars: vanilla
twins.}
\label{fig:pareto}
\end{figure}

\textbf{The price of coverage at the trained exits.} At the two trained
exits below full size,
MLMS retains an edge (per-exit matrix, Appendix~B): 21.89 vs.\
24.38 PPL at 50M and 17.47 vs.\ 19.11 at 100M (best TLM arm vs.\
norm+distill), and it also leads there on benchmarks (36.1 vs.\ 34.1 mean
accuracy at 50M; Appendix~G). This is the expected price
of spreading supervision over twenty prefixes instead of three exits, and
\S\ref{sec:ablations} shows the trade-off is tunable: $\pi$ moves
trained-exit quality and continuum coverage in opposite directions. A budget
midway between the 50M and 100M depths ($k{=}13$) is served by TLM-uniform at
22.6 PPL, nearly matching MLMS's 50M exit (21.89), whereas MLMS at $k{=}13$
collapses to 285 PPL. Any latency target between exits is served by TLM at
near-frontier quality, where MLMS returns a collapsed prefix;
Figure~\ref{fig:pareto} confirms this on \emph{measured} latency rather
than FLOP proxies.

\textbf{Cost.} TLM's advantage is the continuum itself: one run that serves
\emph{any} budget, including budgets not anticipated at training time.
Table~\ref{tab:cost} reports GPU-hours billed by the cluster
(per-run ledger; Appendix~D). The like-for-like comparison
holds distillation off on both sides: the uniform TLM run costs 131 GPU-h
versus 149 for a distillation-free MLMS suite (${\sim}12\%$ cheaper), and
$1.7\times$ less than the vanilla three-model suite (216 GPU-h). MLMS's
cost is higher because each step backpropagates through all $M$ exits (one
shared cascade forward, then a per-exit backward through the shared graph);
TLM instead backpropagates through a single sampled prefix, which is
shallower than the full stack on average, plus the full model. These are
billed GPU-hours for our reimplementation; a different MLMS implementation
could shift the gap. Per delivered operating point this is 6.5 GPU-h (TLM)
versus 50--72 (baselines), though TLM's twenty points include off-frontier
prefixes weaker than the best same-size model, whereas the baselines' three
points all sit on their frontier. When the deployment budgets are few and
known in advance, per-size training is the right tool: three separate
twins cost 216 GPU-h total and give the best per-point quality
(Appendix~B).
Within TLM, the sampler sets cost through the mean sampled depth:
log-uniform $\pi$ is fastest (120 GPU-h), exit-grid slowest (147).

\begin{table}[t]
\centering
\caption{One run, twenty operating points: TLM delivers each at 6.5 GPU-h
versus 50--72 for the suites. Training cost (billed GPU-hours, one A100 per
run, 20B tokens/run); TLM and MLMS rows are single runs, the like-for-like
row holds distillation off on both sides, and TLM's twenty points include
off-frontier prefixes (see text).}
\label{tab:cost}
\small
\begin{tabular}{lccc}
\toprule
Method & GPU-h $\downarrow$ & operating points $\uparrow$ & GPU-h / point $\downarrow$ \\
\midrule
vanilla suite (3 models)      & 216 & 3 & 72 \\
MLMS suite, no distill (1 run) & 149 & 3 & 50 \\
MLMS suite, +distill (1 run)   & 182--185 & 3 & 61--62 \\
\midrule
TLM, uniform $\pi$ (1 run)     & 131 & 20 & 6.5 \\
\bottomrule
\end{tabular}
\end{table}

\subsection{Ablations}
\label{sec:ablations}

\textbf{Prefix sampler $\pi$ is a coverage--exit-quality dial.}
Table~\ref{tab:headline} and the full results matrix (Appendix~B) contain the three-arm $\pi$
ablation ($\lambda=\gamma=1$ fixed, no distillation). Moving $\pi$'s mass
toward the exit depths (log-uniform $\to$ uniform $\to$ exit-grid)
monotonically improves the trained exits (50M: 27.3 $\to$ 26.5 $\to$ 24.4;
100M: 26.6 $\to$ 21.8 $\to$ 19.1). Continuum coverage, by contrast, is best
at the uniform sampler and degrades when mass is concentrated: the exit-grid
arm's off-exit prefixes collapse to MLMS-like values ($k{=}5$: 2{,}693),
confirming that dense stochastic supervision, not the anchor alone, creates
the continuum. AULB is therefore U-shaped in the sampler's mean depth (3.45
log-uniform, \textbf{3.28} uniform, 6.21 exit-grid). Full-size quality is
essentially invariant to $\pi$ (14.99/14.99/14.65); the no-anchor arm
(Appendix~B) confirms the anchor is what preserves the
peak: without it the 200M PPL rises to $22.32$ while the continuum is
unchanged (AULB $3.41$), and a pass-matched variant
(Appendix~B) shows the effect is the full-depth gradient
itself, not the freed compute.

\textbf{The log-uniform arm and the $k{=}1$ prefix.} The log-uniform
sampler $k=\lceil N^u\rceil$ places (near-)zero mass on $k{=}1$, so that
prefix is effectively never trained and collapses (6{,}740 PPL): within
TLM, an unsupervised prefix fails just as MLMS's off-exit prefixes do,
within-run confirmation that supervision, not the nesting, decides whether
a prefix is a language model. Where
it does place mass, log-uniform wins at small prefixes ($k{=}2$: 40.0 vs.\
uniform's 54.7), its intended niche. We therefore use \emph{uniform} $\pi$
as the default: it is the only sampler that trains every prefix, attains the
best AULB, and matches MLMS at full size.

\textbf{Ingredients.} Two further arms isolate the remaining ingredients at
the log-uniform base sampler: one removes the full anchor ($\gamma{=}0$), so
the full model is trained only on the ${\sim}1.7\%$ of steps that sample
$k{=}N$; the other adds MLMS-style stop-gradient distillation into the
sampled prefix ($\alpha_d{=}0.3$). Appendix~B
reports their results; the main-text claims do
not depend on them (every TLM number above is anchor-on and
distillation-free). Appendix~B adds five controls probing
whether each remaining choice is necessary (pass-matched and exit-grid
no-anchor variants, LayerDrop random-depth training with and without
readout calibration, and a dense all-prefix upper bound).

\section{Conclusion}
\label{sec:conclusion}

We asked whether a single training run can deliver a valid language model at
every inference budget. Stochastic prefix supervision with a full anchor, on
an unchanged nested cascade and an identical data stream, delivers that
continuum: with a uniform prefix sampler, all twenty layer prefixes remain
usable language models, perplexity falling smoothly with depth. The
continuum is free at the
peak (full-capacity quality on par with the MLMS suites at ${\sim}12\%$
lower GPU cost) and wins the deployment frontier (LODA rises $1.8\times$;
AULB falls 43--44\%).

\textbf{Limitations.} Our evidence is at the 200M proxy scale on
FineWeb-Edu, with one seed per arm (second seeds for the two headline TLM
arms and paired bootstrap CIs corroborate the main gaps,
Appendix~C); nested and elastic training has been validated
from 100M up to a few billion parameters
\citep{matformer2023,hou2020dynabert,cai2024flextron}, but whether the
peak--continuum separation we report holds at 8B--70B is an open question.
MLMS retains an edge at its two
trained exits below full size, the expected price of coverage. Our budgets
are latency and throughput; on a parameter-count axis the continuum's
advantage shrinks, since prefixes share the embedding and head
(Appendix~E). Serving fixes a depth per request; switching
depth \emph{within} a generation would require truncating or extending the
KV cache across layers, a systems question orthogonal to the training
objective that we leave to future work. Our frontier runs over depth at
fixed width; width, the 3B scale, and speculative decoding are next steps.

\section*{AI-Use Statement}
In this work, we used generative AI tools (AI coding assistants), under human
direction, to aid and polish the writing, to draft and edit sections of the
manuscript, to identify and check relevant literature and references, to
create and refine the figures, and for research ideation and the design and
orchestration of the experiments. We did not use generative AI to generate
synthetic datasets. All experimental results were produced by the logged
training and evaluation runs described in the paper; the authors reviewed all
AI-assisted work and checked every quantitative claim against the recorded
metrics. We take responsibility for the final content of this work, including
text, claims, and artifacts produced with the aid of generative AI.

\section*{Reproducibility Statement}
All runs use the same architecture, data stream, and budget
(\S\ref{sec:setup}); the evaluation protocol (held-out slice, benchmark
configuration, frontier grid) was fixed before the main runs and is
described in \S\ref{sec:setup} and Appendix~G.
The training and evaluation code, including the exact configs and seeds,
will be released publicly upon publication. Per-run
GPU-hour accounting from the cluster's billing records is reported in
Appendix~D. The appendix is provided as a separate supplementary PDF.

\section*{Ethics Statement}
This work trains language models on FineWeb-Edu, a public web-text dataset,
and studies training efficiency; it involves no human subjects and no new
data collection. The energy cost of all experiments is disclosed as
GPU-hours in Appendix~D. More efficient training of
multi-budget model families has broad positive implications for the
environmental cost of deploying language models; we are not aware of
specific dual-use concerns beyond those of language modeling in general.

\section*{Acknowledgements}
This work was supported by a UKRI Future Leaders Fellowship [grant number
G127262].

\bibliography{references}
\bibliographystyle{preprint}

\end{document}